\documentclass[journal]{IEEEtran}
\usepackage{graphicx}
\usepackage{subfigure}
\usepackage{amsmath}
\usepackage{booktabs}
\usepackage{multirow}
\usepackage{footnote}
\usepackage{array}
\usepackage{colortbl}

\ifCLASSINFOpdf
\else
\fi

\begin{document}
%
\title{Deep Optimization Model for Screen Content Image Quality Assessment using Neural Networks }
%
%
%

\author{Xuhao~Jiang,
        Liquan~Shen,
        Guorui~Feng,
        Liangwei~Yu,
        and~Ping An
\thanks{This work is supported by the National Natural Science Foundation of China under grant No.61422111, 61671282 and U1301257 and sponsored by Shanghai Pujiang Program (15pjd015) and Shanghai Shuguang Program (17SG37).}
\thanks{L. Shen is with the Key laboratory of Specialty Fiber Optics and Optical Access Networks, Joint International Research Laboratory of Specialty Fiber Optics and Advanced Communication, Shanghai Institute for Advanced Communication and Data Science, Shanghai University (e-mail: jsslq@163.com).}
\thanks{X. Jiang, G. Feng, L. Yu and P. An are Shanghai Institute for Advanced Communication and Data Science, Shanghai University, Shanghai, China.}}

\maketitle

\begin{abstract}
In this paper, we propose a novel quadratic optimized model based on the deep convolutional neural network (QODCNN) for full-reference and no-reference screen content image (SCI) quality assessment. Unlike traditional CNN methods taking all image patches as training data and using average quality pooling, our model is optimized to obtain a more effective model including three steps. In the first step, an end-to-end deep CNN is trained to preliminarily predict the image visual quality, and batch normalized (BN) layers and $l_2$ regularization are employed to improve the speed and performance of network fitting. For second step, the pretrained model is fine-tuned to achieve better performance under analysis of the raw training data. An adaptive weighting method is proposed in the third step to fuse local quality inspired by the perceptual property of the human visual system (HVS) that the HVS is sensitive to image patches containing texture and edge information. The contribution of our algorithm can be concluded as follows: 1)  Considering the characteristics of SCIs, a deep and valid network architecture is designed for both NR and FR visual quality evaluation of SCIs; 2) with the consideration of correlation between local quality and subjective differential mean opinion score (DMOS), a training data selection method based on effectiveness is proposed to fine-tune the pretrained model; 3) an adaptive pooling approach is employed to fuse patch quality of textual and pictorial regions, whose feature only extracted from distorted images owns strong noise robust and effects on both FR and NR IQA. Experimental results verify that our model outperforms both current no-reference and full-reference image quality assessment methods on the benchmark screen content image quality assessment database (SIQAD). Cross-database evaluation shows high generalization ability and high effectiveness of our model.
\end{abstract}


%
\IEEEpeerreviewmaketitle

\section{Introduction}
%
%
%
%
\IEEEPARstart{N}{owadays} screen content pictures have become quite common in our daily life with the rapid development of multimedia and social network. Numerous consumer applications, such as Facebook, Twitter, remote control and more, involve computer-generalize screen content images (SCIs). Fig. 1(a)-(b) shows two typical images, one is a natural image (NI) and the other is a SCI. There are significant differences between these two images. NIs have rich color and slow color change, while SCIs contain more thin lines, sharp edges and little color variance for massive existence of texts and computer-generated graphics. During acquisition, processing, compression, storage, transmission and reproduction, digital images may introduce various types of distortions, and the visual quality of the images is degraded as a result. Image quality assessment (IQA) aims to objectively evaluate image quality, in order to solve the problem that the human spend much time on judging the image subjective quality. IQA methods can also be used to optimize image processing algorithms. Therefore, IQA plays a very important role in image processing community.
\begin{figure}[t]
\centering
\subfigure[]{
\includegraphics[width=4cm,height=2.6cm]{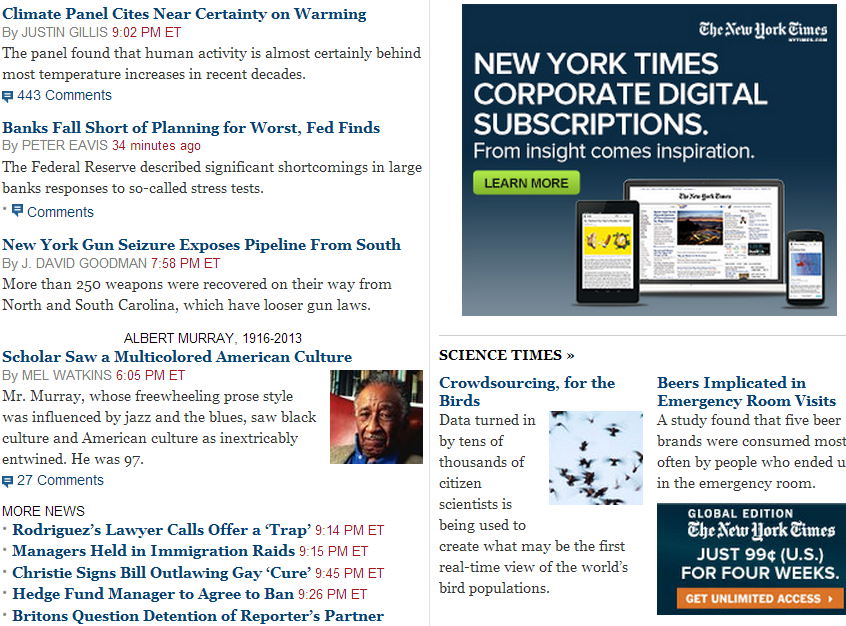}}
\hspace{0in}
\subfigure[]{
\includegraphics[width=4cm,height=2.6cm]{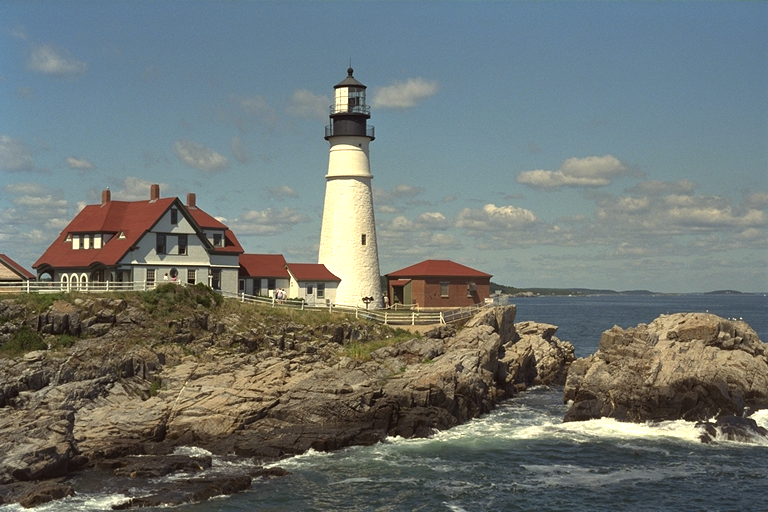}}
\caption{(a) is an example of screen content image and (b) is an example of natural image}
\end{figure}

There are numbers of IQA methods for NIs designed in recent years including full reference (FR), reduced reference (RR) and no reference (NR). Considering the characteristics of the HVS, many FR approaches develop and become highly consistent with subjective quality scores, including structural similarity (SSIM) \cite{wangz.{2004}}, multi-scale SSIM (MS-SSIM) \cite{wangz.{2003}}, information weighted SSIM (IW-SSIM) \cite{wangz.{2011}}, feature similarity (FSIM) \cite{zhangl.{2011}}, visual saliency-based index (VSI) \cite{zhangl.{2014}}, gradient magnitude similarity deviation (GMSD) \cite{xuew.{2014}}, and gradient SSIM (GSIM) \cite{liua.{2012}}. For RR methods, such as \cite{wuj.{2015}}-\cite{mal.{2012}}, only partial information of reference images is used for IQA. However, while evaluating image quality, the reference image or part of its information is unavailable. For NR methods, only the distorted images are employed for IQA. Generally, NR IQA algorithms extract specific features from distorted images and train a regression model with these features and subjective rating by machine learning, such as \cite{saadma.{2012}}-\cite{zhangl.{2015}}.

Most existing IQA approaches devised for NIs are not effective for SCI quality evaluation without taking into account the difference in image content and characteristics between SCIs and NIs. Recognizing this difficulty, others have instead developed a variety of IQA approaches tailored to SCIs. Some representative works \cite{yangh.{2015}}-\cite{fuy.{2018}} of FR IQA have been published and achieves good results. Yang \emph{et al.} propose a FR-IQA model based on SCIs segmentation \cite{yangh.{2015}}. It provides an effective segmentation method to distinguish the pictorial and textual regions, and an activity weighting strategy is employed to fuse the visual quality scores of entire, textual and pictorial regions to the overall quality scores. Based on this segmentation method, structural features based on gradient information and luminance features are extracted for similarity computation to obtain the visual quality of SCI \cite{fangy.{2017}}. In \cite{guk.{2016}}, Gu \emph{et al.} propose a FR metric which mainly relies on simple convolution operators to detect salient areas. Ni \emph{et al.} \cite{niz.{2018}} design a FR metric based on local similarities extracted with Gabor filters in the LMN color space. These FR-IQA methods above achieve a superior performance of SCIs quality evaluation. Numerous NR IQA methods for SCIs are proposed \cite{shaof.{2018}},\cite{Fangy.{2018}}. Shao \emph{et al.} \cite{shaof.{2018}} propose a blind quality predictor for SCI to explore the issue from the perspective of sparse representation. Local sparse representation and global sparse representation are conducted for textual and pictorial regions, respectively. Then, the local and global quality scores are estimated and combined to a total one. Another effective approach of no-reference SCI quality assessment is presented in \cite{Fangy.{2018}}, which obtains an overall quality score through extracting features from the histograms of texture and luminance and training these features based on SVR.

With the development of the CNN, many models \cite{kangl.{2014}}-\cite{bosses.{2018}} have started to build neural networks to process the problem for NI quality assessment, and have achieved superior performance. These methods utilize image patches as a data augmentation, and design special patch-level neural networks for NI IQA. Kang \emph{et al.} \cite{kangl.{2014}} propose a method based on CNN to accurately predict NI quality without reference images. Bosse's method \cite{bosses.{2018}} promotes the CNN to learn the local quality and local weights, and then fuse the local quality to global quality with the local weights. This work mainly considers the relative importance of local quality to the global quality estimation. However, these CNN-based methods still do not consider the special characteristics of the screen content images where textual regions attract more attention than pictorial regions. Zhang \emph{et al.} \cite{zhangy.{2018}} propose a FR-IQA model for SCI taking fusion of textual and pictorial regions into consideration, where the IQAs of pictorial and textual regions are evaluated separately and fused with a region-size-adaptive quality-fusion strategy. Zuo \emph{et al.} propose an NR method using classification models for SCI quality assessment in \cite{zuol.{2016}}. A novel classification network is designed to train the distorted images for getting a practical model, and weightings of texture regions and pictorial regions are determined according to the gradient entropy adapt to the characteristics of the screen content image. Chen \emph{et al.} \cite{Chenj.{2018}} propose naturalization module to transform IQA of NIs into IQA of SCIs. These CNN-based IQA methods for SCIs have their limitations. They divide SCIs into image patches aiming to obtain enough training data, and utilize DMOSs as ground truth. This brings two problems. They are lacking reliable ground truth of image patches and effective strategy of fusing local quality.

In this work, we propose a novel algorithm for both NR and FR IQA of screen content images to solve limitations of the previous methods for screen content images based on patch-level CNN-based models. Unlike traditional patch-level CNN-based methods, our model selects part of all image patches as effective input data whose quality is relatively close to DMOS. For this purpose, a two-steps training strategy is devised. In the first step, the network is trained with all of the image patches to obtain a pretrained model, and predicts the quality scores of the training image patches with the model. Then, the Euclidean distance is employed to evaluate the effectiveness of training image patches, and the pretrained model is fine-tuned with selected image patches to gain a more accurate model. On this basis, an efficient and adaptive weighting method is designed to fuse the visual quality of textual and pictorial regions with considering the effect of the different image patch content. The main contributions of our method are described as follows:

1) Considering the characteristics of SCIs, a deep and valid network architecture is designed for both NR and FR visual quality evaluation of SCIs. Moreover, reference information is extracted by independent layers, which is concatenated with distorted information in the shallow layer for FR-IQA. This confirms that networks learns the feature differences.

2) Considering the connection between the histogram distribution of local quality and DMOSs, the Euclidean distance between local quality and DMOSs is utilized to evaluate the effectiveness of training image patches. A training data selection based on effectiveness is proposed to fine-tune pretrained model for obtaining a higher-performance model.

3) The noise robust index variance of local standard deviation (VLSD) is utilized to distinguish textual and pictorial regions of SCIs, and measure patch weights of two regions. Our proposed adaptive weighting method using VLSD is appropriate to fuse local quality under different types and degrees of distortions.

The rest of this paper is organized as follows. Section II provides a brief review of the related work. In Section III, an effective CNN-based NR-IQA algorithm for SCI is proposed. Section IV shows experimental results and compares performance of the proposed algorithm with the state-of-the-art methods. Finally, conclusions are given in Section V.

\begin{figure}[ht]
\centering
\includegraphics{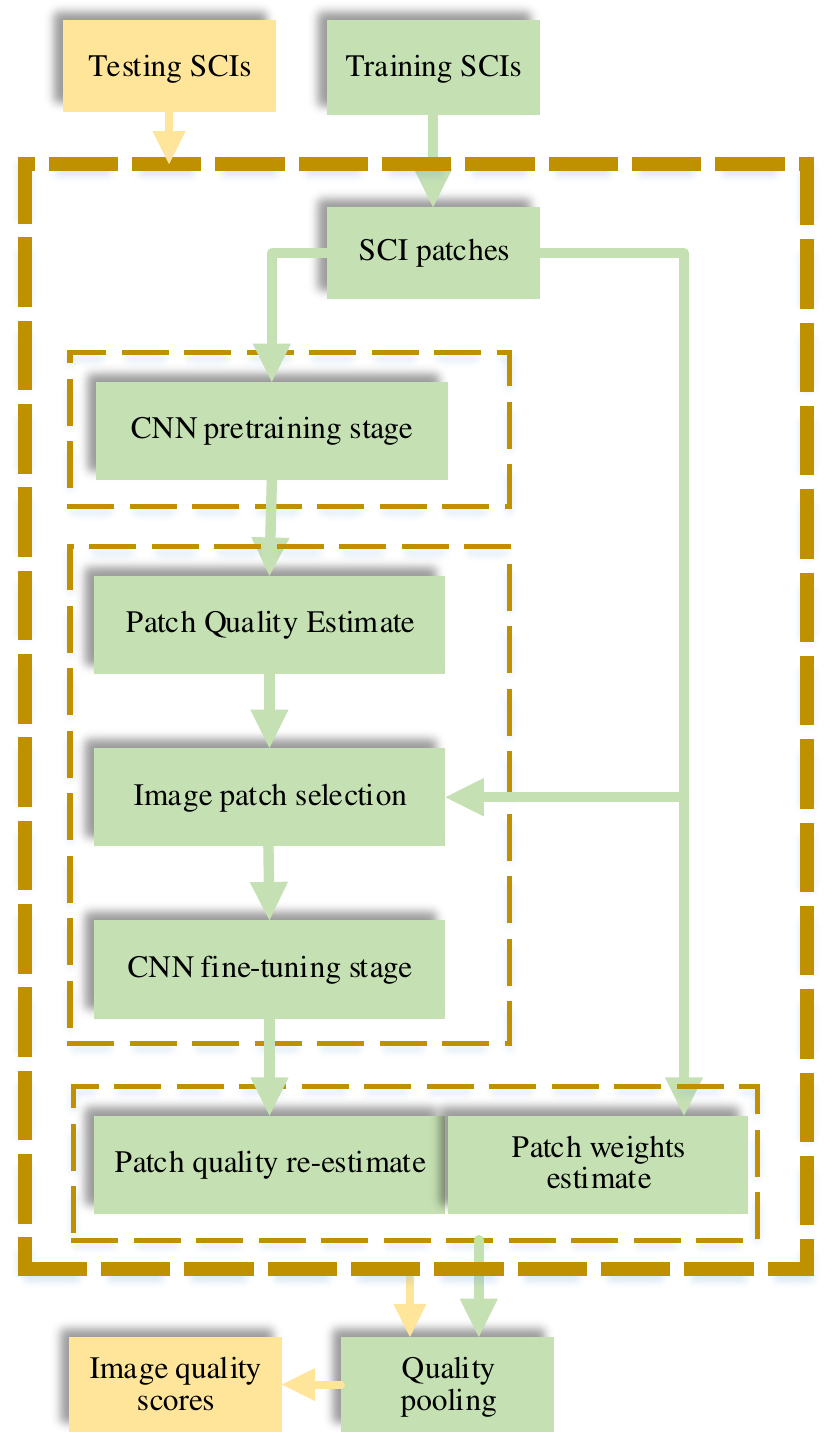}
\caption{Framework of the proposed algorithm.}
\end{figure}

\section{Related Work}
\subsection{Training Patch-level Deep IQA Model with More Reliable Ground-truth}
The traditional CNN-based approaches \cite{kangl.{2014}}, \cite{bosses.{2018}} work with image patches by assigning the subjective DMOS of an image to all patches within it. These approaches suffer from limitation that local quality of image patches within a large image varies even when the distortion is homogeneously applied \cite{wangz.{2004}}. Therefore, some works \cite{bare.{2017}}, \cite{kimjf.{2017}} make use of FR-IQA methods for quality annotation. Kim and Lee \cite{kimjf.{2017}} pretrain the model with the predicted local score of an FR-IQA approach as the ground-truth and fine-tune it with DMOSs. Inspired by this work, Bare \cite{bare.{2017}} devise an accurate deep model which utilizes the FSIM \cite{zhangl.{2011}} to generate training labels of image patches and adopts a deep residual network \cite{hek.{2015}} showing strong ability to extract features in classification and regression tasks. Compared with approaches using DMOSs, utilizing the local score of an FR-IQA model achieves better performance, that benefits network training while each image patch labeled with a more accurate score. Howevere, this brings new problem that the accuracy of the FR-IQA models affects their performance.

Inspired by these works, we observe that the local quality within context of SCIs (e.g., the quality of a $32\times32$ patch within a large image) has great difference for SCIs contain complex content including textual and pictorial regions. However, three high-performance FR-IQA methods for NIs and SCIs are utilized to predict the image patches quality and the performance with an average pooling strategy is poor in Section III-B. The main reason is that local quality of SCIs varies greatly for different characteristics of pictorial and textual regions existing in SCIs. Therefore, the method to train a deep model with the predicted local scores by FR models is not appropriate for SCIs.
\subsection{Salience and Attention for IQA}
Image saliency is one of the most popular topics in computer vision for the characteristics of HVS. This leads to an idea to combine salience models with NR-IQA models. Zhang \emph{et al.} \cite{zhangl.{2014}} devise a VSI FR-IQA model which takes the local saliency from reference and distorted images as feature maps and combines it with similarity from local gradients and local chrominance. Saliency detection is difficult in noisy images that HVS is sensitive to noise. Bosse \emph{et al.} \cite{bosses.{2018}} first propose a learning model to combine saliency with NR-IQA model. In this work, it contains two sub-networks to separately learn local quality and local weight. Then the image visual quality $Q$ is evaluated by weighting the local quality $y_i$ of region $i$ with the corresponding local saliency $w_i$ with
\begin{equation}
  Q=\frac{\sum_i w_iy_i }{\sum_i w_i}
\end{equation}

However, we observe that the method to learn local weight surely improves performance of IQA for NI but has little effect on IQA for SCIs. For NIs, CNN-based model can precisely predict local quality and show a high performance on IQA. For SCIs, CNN-based model does not achieve an expected performance for SCI quality assessment compared with performance of NI quality assessment. Local weight of learning model has high correlation with local quality, and thus the weight prediction will be poor without accurate quality prediction.
\begin{figure*}[ht]
\centering
\includegraphics[width=18cm]{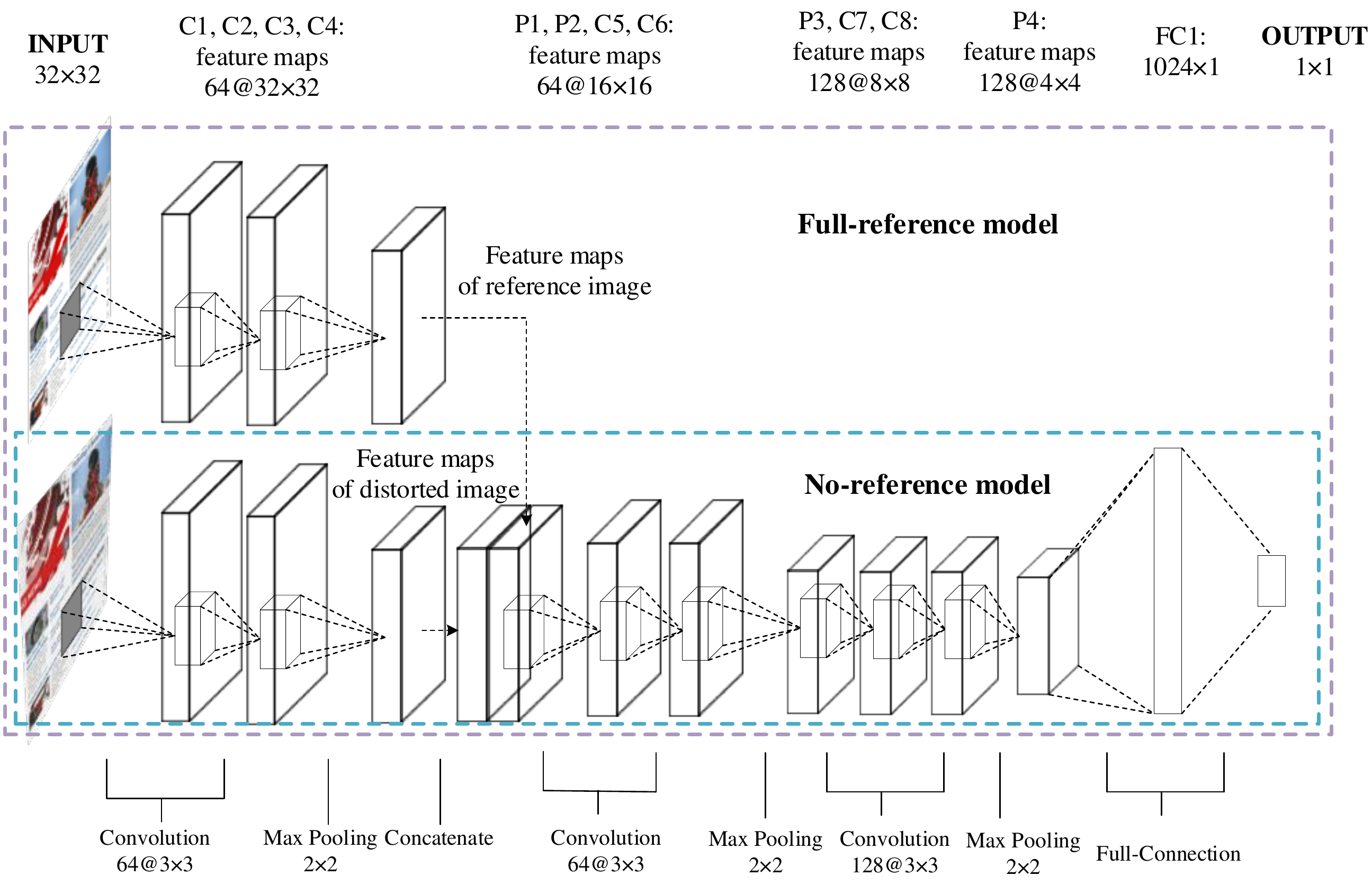}
\caption{An illustration of the architecture of our CNN model.}
\end{figure*}
\section{Proposed Method}
In this section, the proposed QODCNN for SCIs is described in details. As is shown in Fig. 2, QODCNN consists of three sub-steps accomplished by training, fine-tuning and post-processing. Fiestly, the designed CNN is trained with all the image patches to obtain an initial model of SCI visual quality assessment in the first stage, which learns the features of image distortion information and can effectively predict the quality of SCIs. Secondly, with the pre-trained CNN model, quality scores of all the training image patches is predicted and then a data selection is applied according to the predicted scores. Fine-tuning the network aims to gain a more precise and valid model with selected data in the second stage which is as the first optimization. Third, considering the different importances of textual and pictorial regions for IQA of SCIs, the VLSD is designed to distinguish textual and pictorial regions, measure the local weights of image patches and fuse local quality which is the second optimization. Finally, a learning model is obtained to effectively evaluate the visual quality of SCIs.
\subsection{Network Architecture}
\begin{figure*}[t]
\centering
\subfigure[]{
\includegraphics[width=4cm,height=4.06cm]{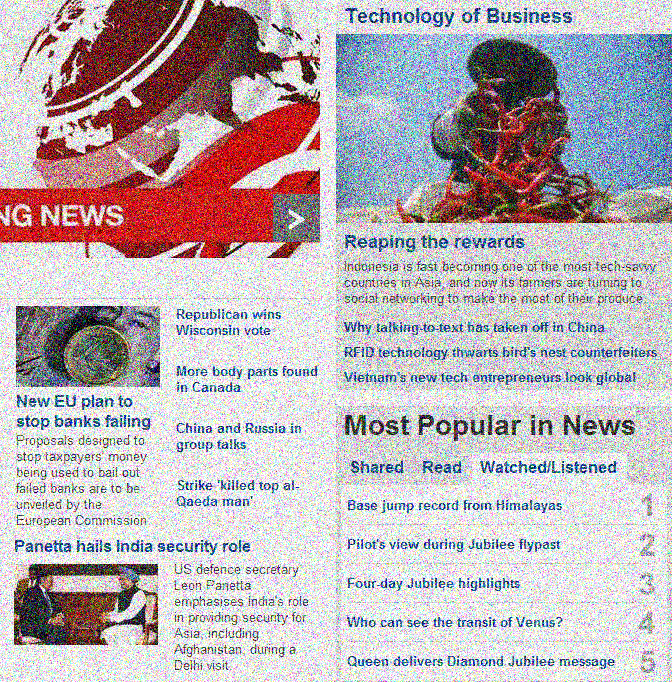}}
\hspace{0in}
\subfigure[]{
\includegraphics[width=4cm,height=4.06cm]{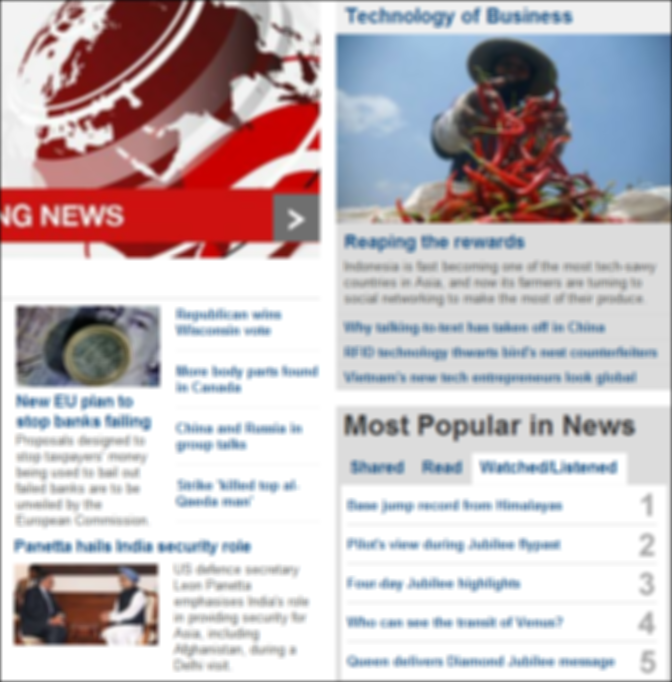}}
\hspace{0in}
\subfigure[]{
\includegraphics[width=4cm,height=4.06cm]{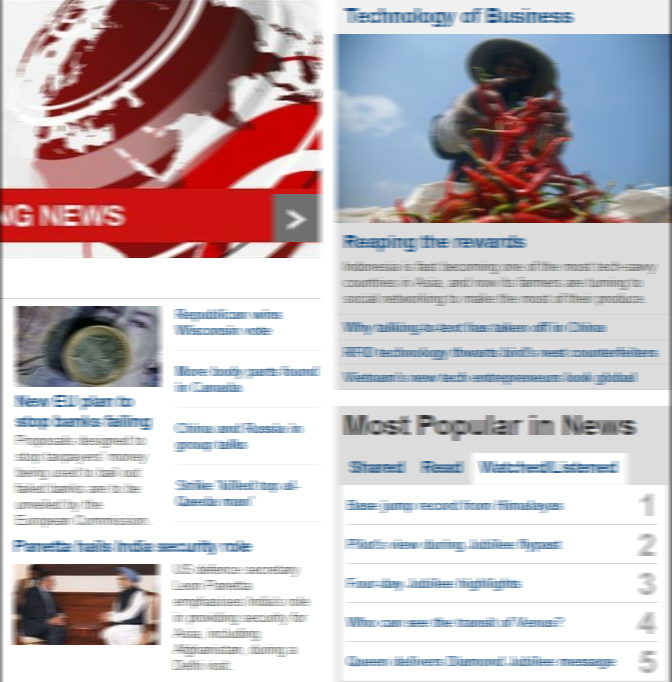}}
\hspace{0in}
\subfigure[]{
\includegraphics[width=4cm,height=4.06cm]{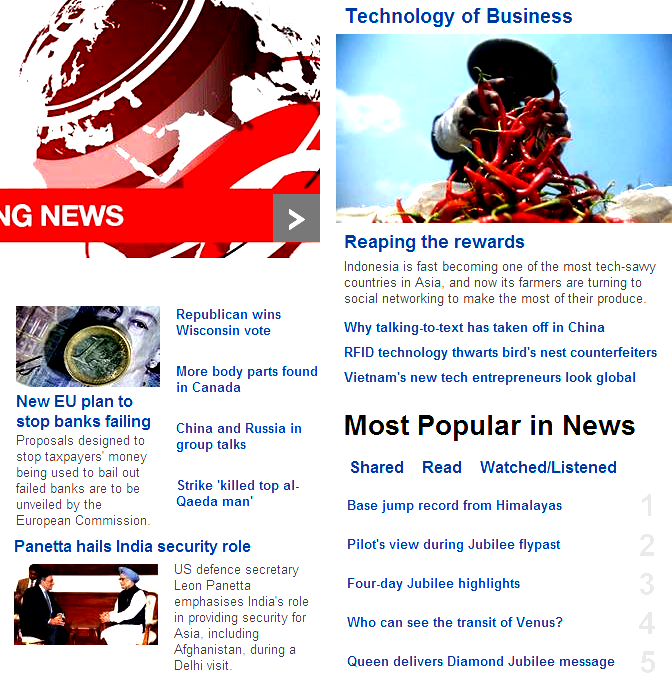}}

\subfigure[]{
\includegraphics[width=4cm]{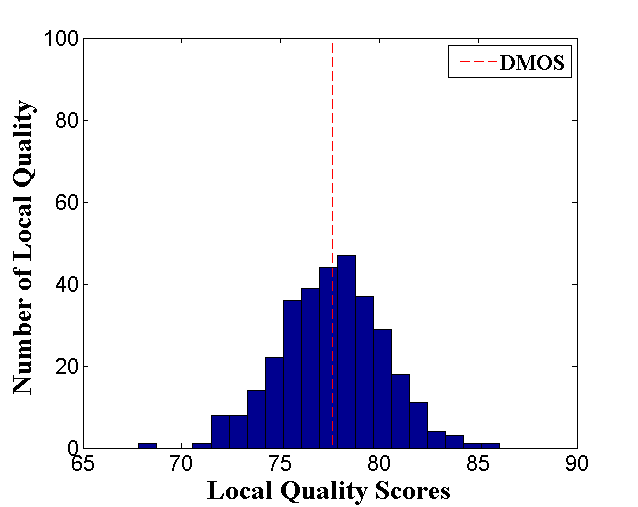}}
\hspace{0in}
\subfigure[]{
\includegraphics[width=4cm]{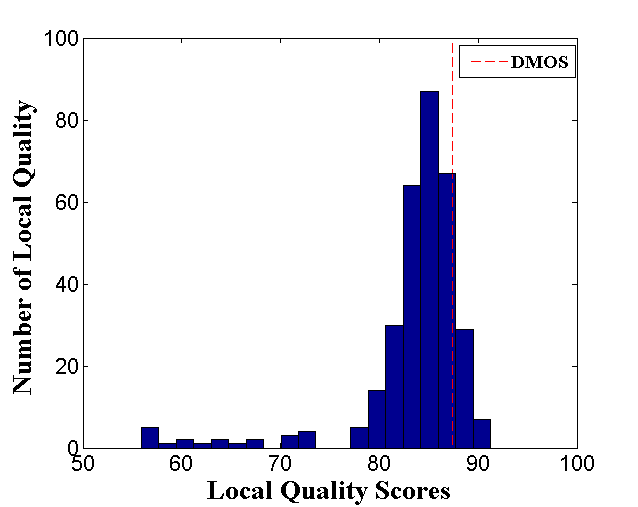}}
\hspace{0in}
\subfigure[]{
\includegraphics[width=4cm]{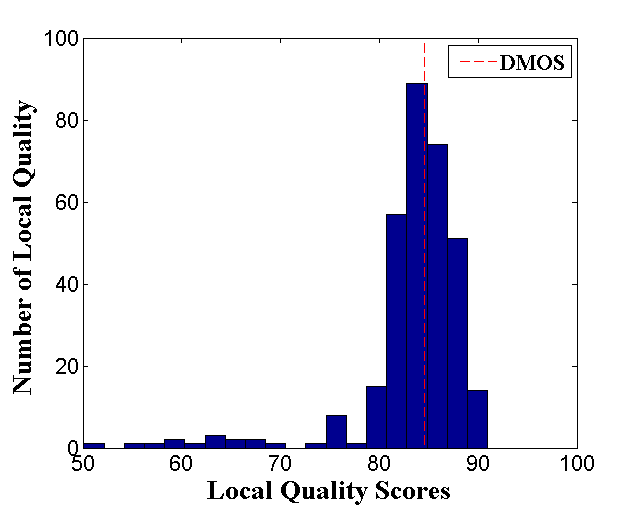}}
\hspace{0in}
\subfigure[]{
\includegraphics[width=4cm]{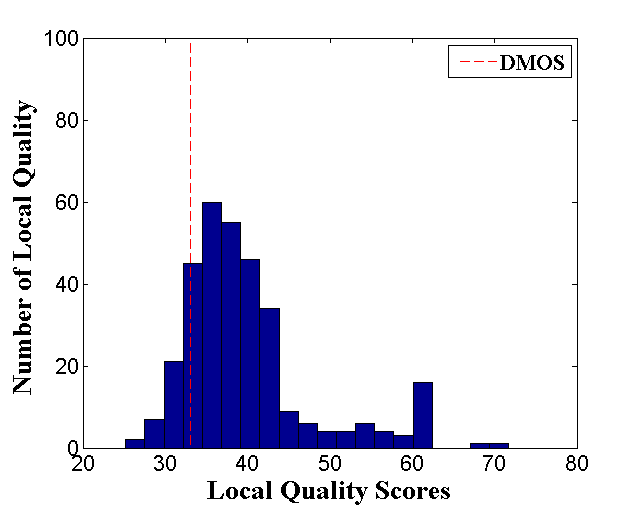}}

\subfigure[]{
\includegraphics[width=4cm,height=4.06cm]{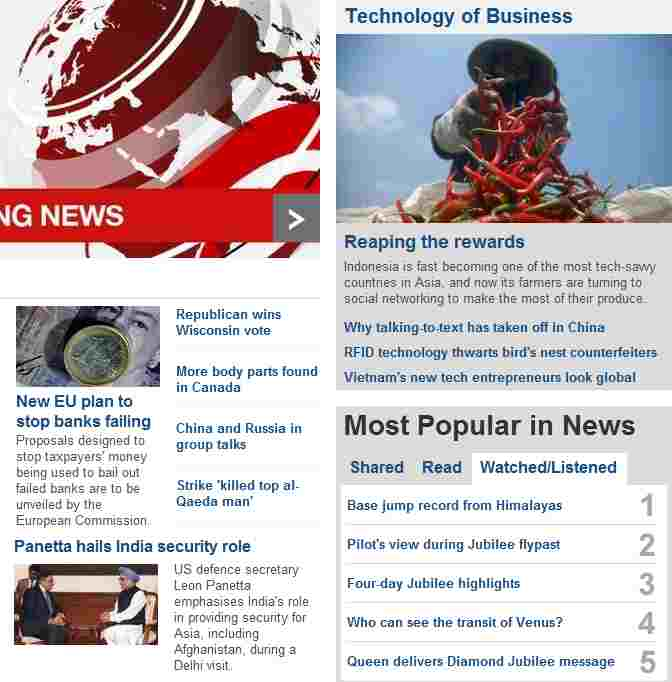}}
\hspace{0in}
\subfigure[]{
\includegraphics[width=4cm,height=4.06cm]{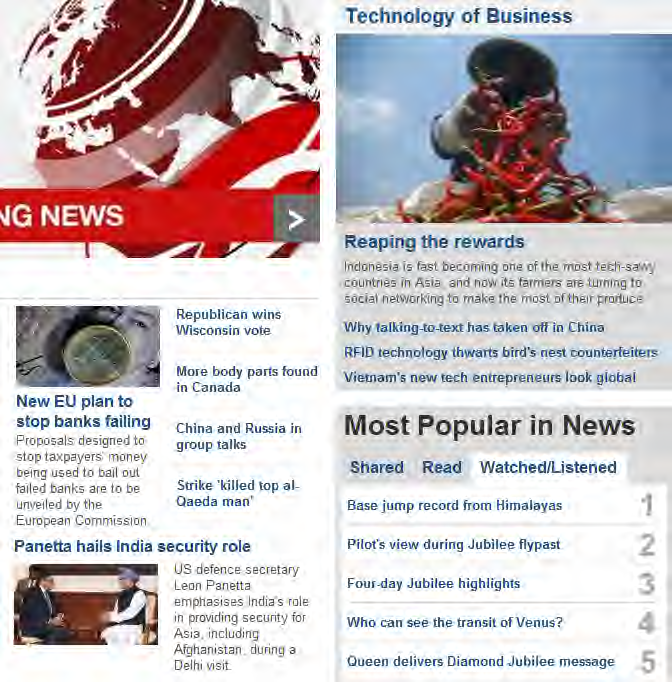}}
\hspace{0in}
\subfigure[]{
\includegraphics[width=4cm,height=4.06cm]{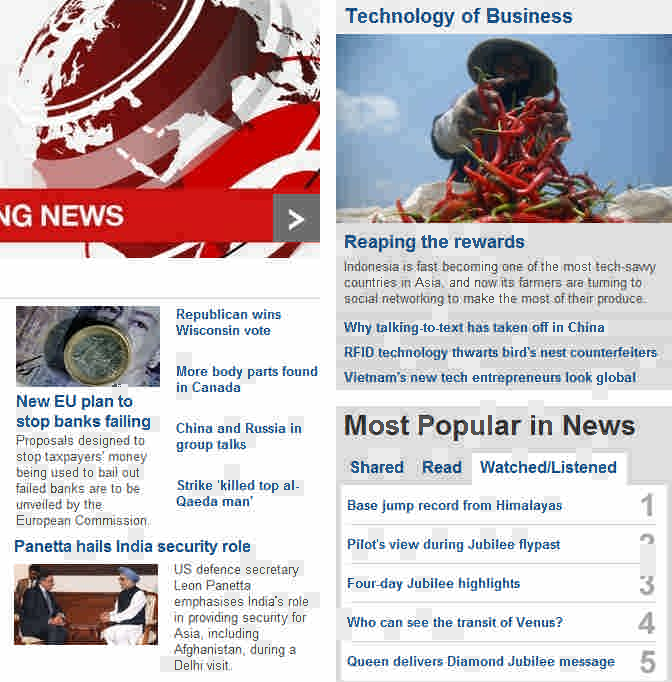}}

\subfigure[]{
\includegraphics[width=4cm]{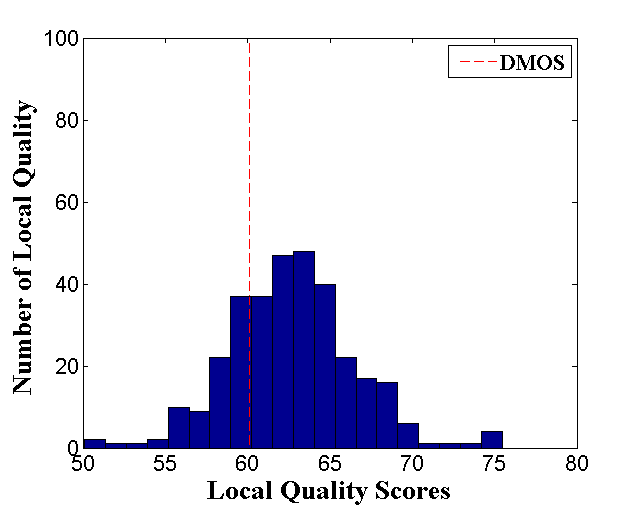}}
\hspace{0in}
\subfigure[]{
\includegraphics[width=4cm]{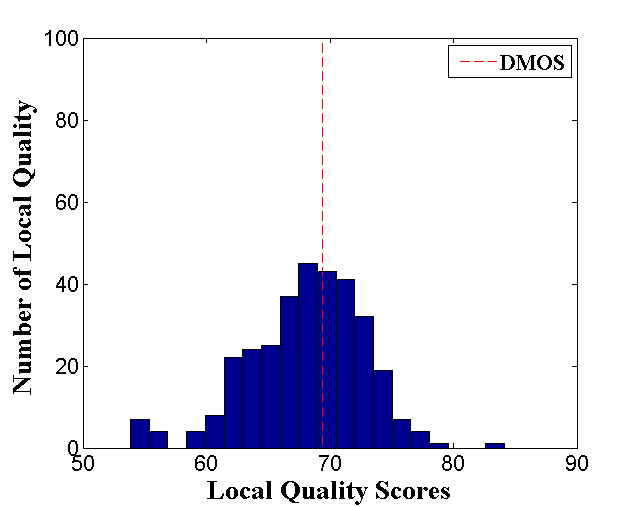}}
\hspace{0in}
\subfigure[]{
\includegraphics[width=4cm]{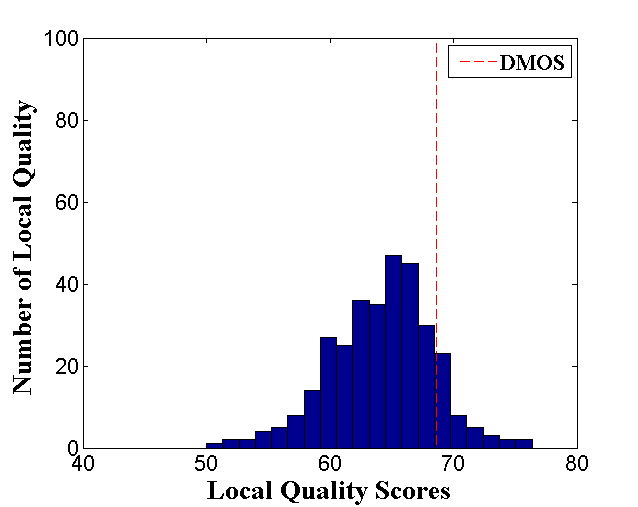}}
\caption{The samples of predicted local quality. The first and third rows show the distorted SCIs; the second and fourth rows show the correspond local quality histograms. (a-d, i-k) are distorted SCIs with the most serious noise of GN, GB, MB, CC, JPEG, JPEG2000, and LSC in SIQAD.}
\end{figure*}
The design of the proposed QODCNN architecture is shown in Fig. 3. The proposed model consists of eight convolutional layers, four max pooling layers, one concatenate layer and two full-connection layers. Each convolutional layer has a $3\times3$ filter with a stride of 1 pixel, and each pooling layer has a $2\times2$ pixel-sized kernel with a stride of 2 pixels. For each convolutional layer, zeros are padded around the border and a BN \cite{ioffes{2015}} layer is added to improve network training performance. The output feature map of the BN layer is calculated by Eq. 2,
\begin{equation}
    y=alpha\times\frac{(Z_j-u)}{std}+beta
\end{equation}
where $Z_j$ is the input feature map of the $j^th$ batch normalized layer and $y$ is the output. $u$ and $std$ respectively denote the mean and variance of the input map. $alpha$ and $beta$ are two parameters updated in training. The rectified linear unit (ReLU) \cite{nairv.{2010}} as activation function is added after the normalized layers. Feature maps extracted by convolutional layers and pooling layers are named, and the precise configurations are listed in Fig. 3.

For FR-IQA, this model extracts the feature maps of reference image patches and distorted image patches, and fuses these maps with a concatenate layer in shallow layer of network.  These fused feature maps are regressed by the remaining network layers. For NR-IQA, the branch of extracting feature maps of reference images is abolished, and thus the adjusted model extracts features only from the distorted images. The $l_1-norm$ loss function for both NR and FR models is defined as follows,
\begin{equation}
  l_1=\frac{1}{N}\sum_{i=1}^N|\hat{q}-q|
\end{equation}
where $N$ is the image patch number of an input mini-batch , $\hat{q}$ is the network output of an input image patch, and $q$ is the ground-truth of the input image patch.

The CNN parameters are learned end-to-end by minimizing the sum of loss function and $l_2$ regularization for the predicted quality, on all training tuples:
\begin{equation}
  min\{\frac{1}{N}\sum_{i=1}^N|\hat{q}-q|+\frac{\alpha}{2N}\sum w^2\}
\end{equation}
where $\alpha$ represents the penalty factor and $w$ is the weight of CNN model.

In our model, combination of two $3\times3$ convolutional layers and one pooling layers is employed for owning a larger view to extracting features with less data. Considering SCIs contain lots of edge and gradient information, the max pooling layer is applied to capture texture changes degraded by noise. For FR-IQA, reference information is extracted by independent layers and concatenated with distorted information in the shallow layer. This confirms that networks learns the differences between features extracted from distorted and reference SCIs rather than the differences between distorted and reference SCIs. Compared with features in deep layer, features in shallow layer retain a large amount of original information with less information loss. In addition, BN layers and $l_2$ regularization significantly improve the speed of the model's regression and the performance of fitting. Most existing models adopt $l_2-norm$ loss for NI quality assessment. However, considering the big statistics differences between image patches of SCIs, $l_1-norm$ loss is applied to reduce the impact of some abnormal image patches.
\subsection{A Two-steps Training Strategy}
Most existing patch-level CNN-based models all face a problem that local quality of a large image is labeled with an inaccurate quality score. This problem is more serious when employing a patch-level CNN model to predict visual quality of SCIs. Compared with NIs, SCIs contain more complex content consisting of pictorial and textual regions. Some approaches \cite{bare.{2017}}, \cite{kimjf.{2017}} utilize FR-IQA methods to predict local quality of natural scene image. However, it has limited effect for SCI visual quality evaluation. Here, SSIM \cite{wangz.{2004}}, FSIM \cite{zhangl.{2011}} and SQMS \cite{guk.{2016}} are employed to test the performance where the three FR-IQA methods predict local quality of SCIs and fusing local quality with average pooling is applied to obtain image quality on SIQAD database \cite{yangh.{2015}}. From Table I, it can be observed that these methods illustrate poor performance due to existing big difference between image patches of SCIs. Therefore, we consider solving this problem from SCI patches itself, not from the correspond labels.
\begin{table}[!htbp]
\centering
\caption{Patch-Level Performance of Three FR-IQA Methods}\label{tab:aStrangeTable}
\begin{tabular}{ccccc}
\toprule[1pt]
\textbf{\emph{Index}}& PLCC& SRCC &RMSE\\
\midrule[0.5pt]
SSIM \cite{wangz.{2004}}& 0.7630& 0.7602 & 10.8837\\
FSIM \cite{zhangl.{2011}}& 0.8295& 0.8328 & 9.2777\\
SQMS \cite{guk.{2016}}& 0.8104& 0.8156 & 10.2790\\
\bottomrule[1pt]
\end{tabular}
\end{table}

In our model, CNN trained with more efficient data (TMED) are considered to solve this problem including two training steps. In the first step, an initial model is obtained by training neural network of Section III A with all the image patches for SCI quality assessment. To test effectiveness of the pretrained model, this model predicts all the training image patches and the corresponding distribution maps of these predicted patch scores are provided in Fig. 4. As shown in Fig. 4, the distorted images of seven different distortion types and the corresponding local quality histograms are listed. For gaussian noise (GN), gaussian blur (GB), motion blur (MB), contrast change (CC), JPEG compression (JC), JPEG2000 compression (J2C) and layer segmentation-backed coding (LSC) distortions, the most serious noise is used to analysis as typical examples. These histograms of local quality show that predicted local quality of training image patches is distributed around the DMOSs. This also verifies the view mentioned in the introduction that the local quality of a large image varies. It can be noted the local quality scores of some image patches are far away from the DMOSs which damages the performance of the learning model.

Inspired by histograms of local quality, a training data selection is more reasonable and benefits the deep model learning, since CNN can be learned better with training data labeled with precise ground-truth. Data selection abandons those image patches whose local quality deviates from ground-truth and selects those local quality closes to DMOSs. Therefore, the Euclidean distance is employed to evaluate the effectiveness of training image patches and calculated as follows,
\begin{equation}
  E=|\hat{q}-q|
\end{equation}
where $\hat{q}$ is the predicted score of an image patch, $q$ is the ground-truth of the image patch and $E$ denotes the effectiveness index. In order to maintain enough information of each image, image patches are selected from each image with a fixed ratio. The ratio is computed by
\begin{equation}
  P=\frac{N_s}{N}~~~~(E\leq T)
\end{equation}
where $P$ presents the data selection ratio, $N_s$ and $N$ are the number of selected image patches and all image patches of a SCI, and $T$ denotes an adaptive threshold. The pre-trained model is fine-tuned with selected training data to obtain a more effective and higher-performance model in the second step, .
\subsection{Pooling with A Novel Weighting Method}
The pooling-by-average approach is used to calculate global quality with local quality in most patch-level CNN-based models \cite{kangl.{2014}}, \cite{bare.{2017}}. However, the average pooling local quality estimates does not consider the effect of spatially varying perceptual relevance of local quality. Especially for SCIs, HVS is sensitive to edge information meaning that textual regions owns higher weights than pictorial regions for IQA. It is very difficult to distinguish these two regions only from distorted images. Fang's method \cite{fangy.{2017}} obtains global quality by weighting local quality with gradient entropy. Using gradient entropy to distinguish textual and pictorial regions is useful for FR-IQA model, but it is difficult for NR-IQA since entropy is sensitive to noise.

\begin{figure}[ht]
\centering
\includegraphics{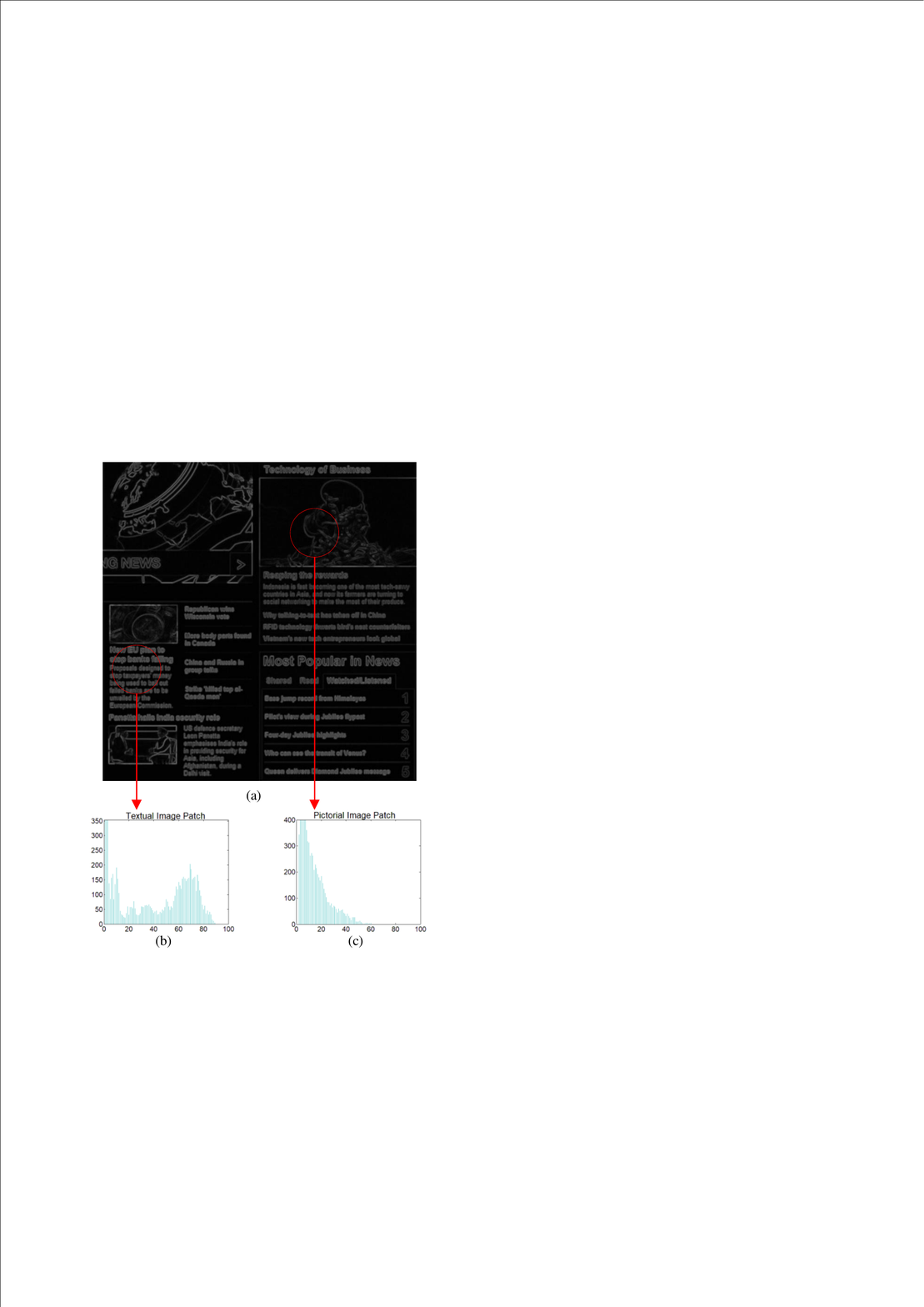}
\caption{(a) is the map with a smooth processing of a SCI distorted by Gaussian noise; (b) and (c) are of different pixel distributions of pictorial region and textual region in the smoothed image.}
\end{figure}

In our model, weighting local quality with VLSD (WLQVLSD) is first proposed for SCI quality assessment. Fig. 5(a) depicts map of a typical gaussian distorted SCI with a smooth processing named LSD, and Figs. 5(b) and (c) depict two histograms of the textual region and pictorial region. LSD is applied on each image to indicate the structural complexity and reduce the impact of noise. The value of output feature map is calculated as,
\begin{equation}
  LSD(i,j)=\sqrt{\sum_{k=-K}^K\sum_{l=-L}^Lw_{k,l}(I_{k,l}(i,j)-\mu(i,j))^2}
\end{equation}
where $w={w_{(k,l)} |k=-K,...,K,l=-L,...,L}$ is a 2D circularly-symmetric Gaussian weighting function, $I_{k,l}(i,j)$ is the pixel point in the image, and $\mu(i,j)$ is the mean value of within a $K\times L$ local window centered at $(i,j)$, $LSD(i,j)$ is the output pixel of the corresponding position. In our implementation, $K=L=3$. As shown in Fig. 5(a), the noise is weakened after smoothing, and plenty of thin lines are left on the image which is beneficial for the distinguishing between pictorial and textual regions.

Second, two typical regions are marked in Fig. 5(a). It can be seen that the histogram of the textual region (b) is relatively scattered compared to the pictorial region (c) whose histogram is relatively concentrated. Considering the differences of the histogram distribution, the variance of LSD is taken as the feature to describe the contents of image. The value of variance is calculated by,
\begin{equation}
VLSD=\frac{1}{N}\sum_{n=1}^N(LSD(i,j)-\hat{LSD})^2
\end{equation}
where $N$ is number of pixels in the image patch, $\hat{LSD}$ is the mean value of local deviation map and $LSD(i,j)$ is the pixel point in the local deviation map which are calculated in Eq. 7. The reference SCI and the corresponding VLSD maps of three distortion types (GN, CC and JPEG) are shown in Fig. 6. To demonstrate the noise robustness of VLSD, the distorted SCIs with the slightest and most serious noise are as examples. As seen in Fig .6, two typical areas are marked with two different color boxes where yellow boxes represent pictorial regions and blue boxes represent textual regions. It can be observed that textual regions of SCI obtains bigger value compared with pictorial regions of SCI in all VLSD maps. This shows VLSD can effectively distinguish pictorial and textual regions, and measure local weight of SCIs. Moreover, VLSD also owns strong noise robust ability. Compared with gradient entropy, the VLSD can better distinguish textual regions and pictorial regions, and is robust to distortion types and intensity. Thus the VLSD is employed to measure the importance of local regions in a large image.

\begin{figure}[ht]
\centering
\includegraphics[width=8cm]{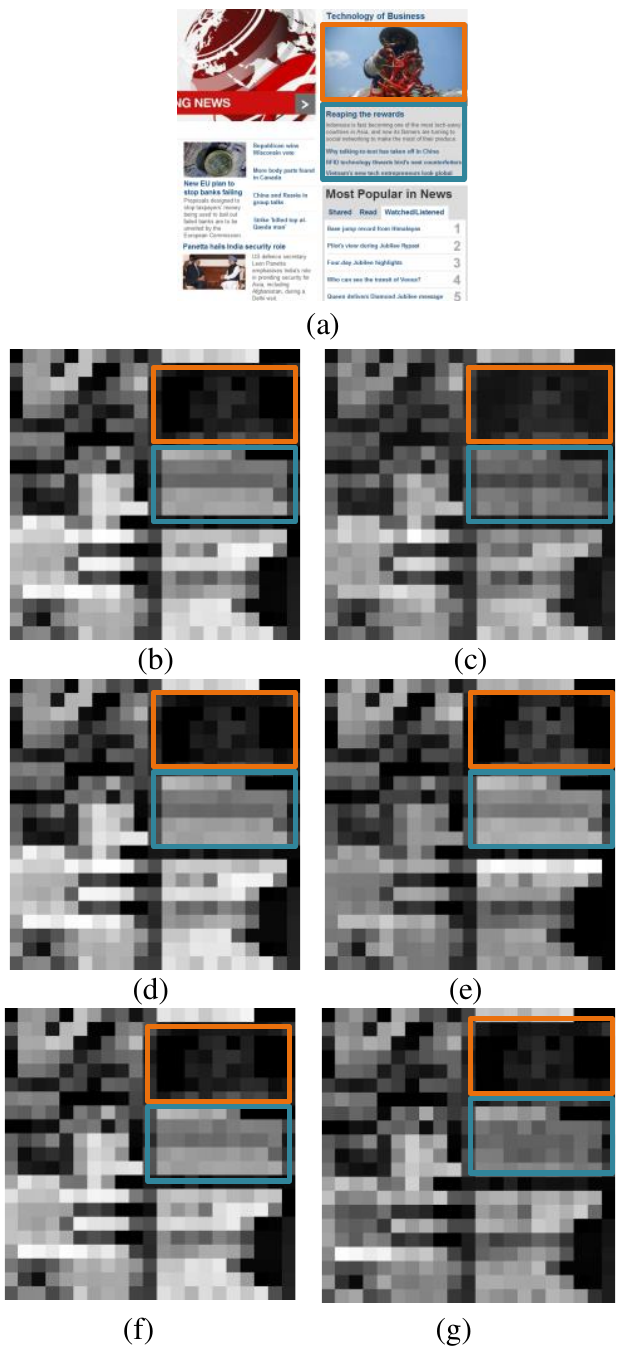}
\caption{The visual samples of VLSD maps. (a) is a reference SCI and (b-g) are VLSD maps of six distorted SCIs. (b, c), (d, e) and (f, g) are VLSD map pairs with the slightest and most serious noise of GN, CC and JPEG in SIQAD.}
\end{figure}




Finally, scores predicted by the fine-tuned CNN and the corresponding VLSD of the image patches are obtained. A weighting method is applied to fuse quality of textual and pictorial regions which is calculated as,
\begin{equation}
  S=\frac{\sum_{n=1}^N s_n\times VLSD_n }{\sum_{n=1}^N VLSD_n}
\end{equation}
where $s_n$ and $VLSD_n$ are the score of the $n^{th}$ patch and its variance value calculated based on Eq. 8, $N$ is the number of the patches of the test image, $S$ is the final score of the test image.
\subsection{Training}
Before training our model, a grayscale processing and a data augmentation are applied by dividing large color images into gray image patches with size $32\times32$. The Tensorflow is used as the training toolbox, and two databases \cite{yangh.{2015}}, \cite{niz.{2017}} are used to train and test our model. Both pre-training and fine-tuning steps adopt the Adam optimization algorithm \cite{kingma.{2014}} with a mini-batch of 64, and employ DMOSs as ground-truth of training. The penalty factor $\alpha$ of $l_2$ regularization is $1\times 10^{-5}$. In the pre-training stage, the learning rate is changed from $1\times 10^{-4}$ to $1\times 10^{-13}$ at the interval of ten epochs. For fine-tuning, we fine-tune the pre-trained model with the same learning rate conditions. After training for two hundred epochs, the final model is obtained to predict visual quality.

\section{Experimental Results}
\subsection{Database and Evaluation Methodology}
To verify the effectiveness of the proposed QODCNN, two screen content image databases SIQAD \cite{yangh.{2015}} and SCID \cite{niz.{2017}} are used to conduct the comparison experiments. The SIQAD database has 20 reference screen content images and 980 distorted screen content images which contain seven types of distortion including GN, GB, MB, CC, JC, J2C and LSC, and each is with seven levels of distortions. The SCID database is used for cross-database experiments. Six distortion types (GN, GB, MB, CC, JC, and J2C) at five different levels is considered that are common in the two databases. This leaves us 1200 test SCIs in SCID.

In most cases, three typical performance evaluation criteria are adopted to evaluate the performance of IQA algorithms, the Pearson Linear Correlation Coefficient (PLCC), Spearman’s Rank-order Correlation Coefficient (SRCC) and Root Mean Square Error (RMSE). The values of PLCC and SRCC is closer to 1, and the values of RMSE is smaller, indicating that the algorithm is more accurate. Given the $i^{th}$ image in the database (with $N$ images in total), $o_i$ and $s_i$ are the objective and subjective scores, $\bar{o}$ and $\bar{s}$ are the mean values of $o_i$ and $s_i$, $e_i$ is the difference between the subjective and objective results. PLCC, SRCC and RMSE are defined as follows,
\begin{equation}
  PLCC=\frac{\sum_{i=1}^N(o_i-\bar{o})(s_i-\bar{s})}{\sqrt{\sum_{i=1}^N(o_i-\bar{o})}\times\sum_{i=1}^N{(s_i-\bar{s})}}
\end{equation}
\begin{equation}
  SRCC=1-\frac{6\sum_{i=1}^Ne_i^2}{N(N^2-1)}
\end{equation}
\begin{equation}
  RMSE=\sqrt{\frac{\sum_{i=1}^N(o_n-s_n)^2}{N}}
\end{equation}
A five-parameter mapping function [50] is employed to nonlinearly regress the quality scores into a common space as follows,
\begin{equation}
  f(x)=\beta_1(\frac{1}{2}-\frac{1}{1+exp(\beta_2(x-\beta_3))})+\beta_4x+\beta_5
\end{equation}
where ($\beta_1,...,\beta_5$) are parameters to be compute with a curve fitting process.
\subsection{Performance Evaluation on SIQAD}
For performance evaluation, the proposed QODCNN model is trained on database SIQAD. To distinguish the NR and FR models, the NR model is named as QODCNN-NR and the FR model is named as QODCNN-FR. For SIQAD, 980 distorted SCIs are randomly divided into two subsets according to the image content. Training set contains 784 distorted SCIs associated to 16 reference images (80$\%$ data for training) and the rest 196 distorted SCIs associated to 4 reference images are used as testing set (20$\%$ data for testing). The training-testing sets partition are randomly repeated 10 times, and the average performance of ten experiments is calculated as the overall performance.

In the second stage of our proposed QODCNN model, the proportion of data selection influences the performance of the fine-tuned model. Here, considering the difference of local quality within a large image, 10$\%$ to 80$\%$ of image patches of each image in training set of the second stage are selected by adjusting the threshold of Eq. 5 for training, and all of the image patches in testing set are utilized to evaluate the performance. Experiments are repeated for different training-testing sets of the first stage and different proportion of training data. For different proportion of training data, the average performance of ten results is calculated as the overall performance. The best performance is obtained when 70$\%$ of training data is used to fine-tune pre-trained model, which can be seen in Table V and Fig. 7 in Section IV-D-1. Thus, proportion of 70$\%$ is applied for experiments on SIQAD and cross-database experiments.

\begin{table}[t]
\centering
\caption{Experimental results of proposed and other existing FR and NR methods on SIQAD database}\label{tab:aStrangeTable}
\begin{tabular}{lllll}
\toprule[1pt]
~&\textbf{\emph{Method}} &PLCC&SRCC&RMSE\\
\midrule[0.5pt]
\multirow{9}{*}{\centering{\rotatebox{90}{Full-Reference}}} &
  PSNR &0.5869&0.5608&11.5859\\
~&SSIM [1]&0.7561&0.7566&9.3676\\
~&GMSD [6]&0.7259&0.7305&9.4684\\
~&SPQA [17]&0.8584&0.8416&7.3421\\
~&ESIM [18]&0.8788&0.8632&6.8310\\
~&SQMS [20]&0.8872&0.8803&6.6039\\
~&GFM [21]&0.8828&0.8735&6.7234\\
~&SFUW [19]&0.8910&0.8800&6.4990\\
~&MDOGS [22]&0.8839&0.8822&6.6951\\
~&CNN-SQE [33]&\textbf{0.9040}&\textbf{0.8940}&\textbf{6.1150}\\
~&QODCNN-FR&\textbf{0.9142}&\textbf{0.9066}&\textbf{5.8015}\\
\midrule[0.5pt]
\multirow{8}{*}{\centering{\rotatebox{90}{No-Reference}}} &
  BLINDS-II [11]&0.7255&0.6813&9.4991\\
~&BRISQUE [12]&0.7708&0.7237&8.1342\\
~&BLIQUP-SCI [23]&0.7705&0.7990&10.0213\\
~&NRLT [24]& 0.8442&0.8202&7.5957\\
~&CNN-Kang [25]& 0.8487&0.8091&7.4472\\
~&WaDIQaM-NR [32]&0.8594&0.8522&7.0570\\
~&PICNN [35]&\textbf{0.896}&\textbf{0.897}&\textbf{6.790}\\
~&QODCNN-NR&\textbf{0.9008}&\textbf{0.8888}&\textbf{6.2258}\\
\bottomrule[1pt]
\end{tabular}
\end{table}

\begin{table*}[!t]
\newcommand{\tabincell}[2]{\begin{tabular}{@{}#1@{}}#2\end{tabular}}
\addtolength{\tabcolsep}{1.5pt}
\renewcommand{\arraystretch}{1}
\caption{Experimental Results of Our Proposed Models and Other State-of-the-art FR Models on Different Distortion Types on SIQAD.}\label{tab:aStrangeTable}
\centering
\scriptsize
\renewcommand{\multirowsetup}{\centering}
\begin{tabular}{ccccccccccccc}
\toprule[1pt]
\multirow{2}{*}{\tabincell{c}{\textbf{\emph{Method}}}}&\multirow{2}{*}{\tabincell{c}{Distortions}}&\multirow{2}{*}{\tabincell{c}{PSNR}}
&\multirow{2}{*}{\tabincell{c}{SSIM}}
&\multirow{2}{*}{\tabincell{c}{GMSD}}
&\multirow{2}{*}{\tabincell{c}{SPQA}}
&\multirow{2}{*}{\tabincell{c}{ESIM}}
&\multirow{2}{*}{\tabincell{c}{SQMS}}
&\multirow{2}{*}{\tabincell{c}{SFUW}}
&\multirow{2}{*}{\tabincell{c}{MDOGS}}
&\multirow{2}{*}{\tabincell{c}{CNN-SQE}}
&\multirow{2}{*}{\tabincell{c}{QODCNN-NR}}
&\multirow{2}{*}{\tabincell{c}{QODCNN-FR}}\cr\\
\cmidrule[0.5pt]{1-13}
\centering{\multirow{8}{*}{PLCC}}
&GN&\textbf{0.905}&0.881&0.899&0.892&0.899&0.900&0.887&0.898&-&\textbf{0.913}&\textbf{0.918}\\
&GB&0.860&0.901&0.910&0.906&\textbf{0.923}&0.912&\textbf{0.923}&0.920&-&\textbf{0.925}&\textbf{0.934}\\
&MB&0.704&0.806&0.844&0.831&\textbf{0.889}&0.867&0.878&0.842&-&\textbf{0.889}&\textbf{0.907}\\
&CC&0.753&0.744&0.783&0.799&0.764&0.803&\textbf{0.829}&0.801&-&\textbf{0.837}&\textbf{0.866}\\
&JC&0.770&0.749&0.775&0.770&\textbf{0.800}&0.786&0.757&0.789&-&\textbf{0.830}&\textbf{0.848}\\
&J2C&0.789&0.775&\textbf{0.851}&0.825&0.789&0.826&0.815&\textbf{0.861}&-&0.818&\textbf{0.857}\\
&LSC&0.781&0.731&\textbf{0.856}&0.796&0.792&0.813&0.759&0.832&-&\textbf{0.867}&\textbf{0.897}\\
&Overall&0.587&0.756&0.726&0.858&0.879&0.887&0.891&0.884&\textbf{0.904}&\textbf{0.901}&\textbf{0.914}\\ \cmidrule[0.5pt]{1-13}

\centering{\multirow{8}{*}{SRCC}}
&GN&0.879&0.870&0.886&0.882&0.876&0.886&0.869&0.888&\textbf{0.893}&\textbf{0.905}&\textbf{0.907}\\
&GB&0.858&0.892&0.912&0.902&\textbf{0.924}&0.915&0.917&0.919&\textbf{0.924}&0.916&\textbf{0.921}\\
&MB&0.713&0.804&0.844&0.826&\textbf{0.894}&0.869&0.874&0.835&\textbf{0.904}&0.871&\textbf{0.895}\\
&CC&0.683&0.641&0.544&0.615&0.611&0.695&\textbf{0.722}&0.664&0.665&\textbf{0.700}&\textbf{0.778}\\
&JC&0.757&0.758&0.771&0.767&0.799&0.789&0.750&0.786&\textbf{0.847}&\textbf{0.815}&\textbf{0.829}\\
&J2C&0.775&0.760&0.844&0.815&0.783&0.819&0.812&\textbf{0.862}&\textbf{0.862}&0.795&\textbf{0.835}\\
&LSC&0.793&0.737&0.859&0.800&0.796&0.829&0.754&0.851&\textbf{0.887}&\textbf{0.882}&\textbf{0.898}\\
&Overall&0.561&0.757&0.731&0.842&0.863&0.880&0.880&0.882&\textbf{0.894}&\textbf{0.889}&\textbf{0.907}\\ \cmidrule[0.5pt]{1-13}

\centering{\multirow{8}{*}{RMSE}}
&GN&\textbf{6.338}&7.068&6.521&6.739&6.827&6.921&6.876&6.558&-&\textbf{6.150}&\textbf{5.963}\\
&GB&7.738&6.570&6.310&6.430&5.827&6.611&\textbf{5.592}&5.964&-&\textbf{5.772}&\textbf{5.454}\\
&MB&9.229&7.697&6.982&7.222&\textbf{5.964}&7.204&6.236&7.012&-&\textbf{5.762}&\textbf{5.251}\\
&CC&8.282&8.412&7.828&7.618&8.114&7.743&\textbf{7.048}&7.528&-&\textbf{6.939}&\textbf{6.381}\\
&JC&6.000&6.230&5.941&6.000&\textbf{5.640}&5.983&6.143&5.779&-&\textbf{5.460}&\textbf{5.141}\\
&J2C&6.382&6.591&\textbf{5.459}&5.871&6.388&6.050&6.023&\textbf{5.293}&-&6.000&\textbf{5.286}\\
&LSC&5.330&5.825&\textbf{4.411}&5.166&5.215&5.104&5.555&4.738&-&\textbf{4.338}&\textbf{3.857}\\
&Overall&11.590&9.368&9.642&7.342&6.831&7.297&6.499&6.695&\textbf{6.115}&\textbf{6.226}&\textbf{5.801}\\
\bottomrule[1pt]
\end{tabular}
\end{table*}

\begin{table*}[!t]
\newcommand{\tabincell}[2]{\begin{tabular}{@{}#1@{}}#2\end{tabular}}
\addtolength{\tabcolsep}{1.5pt}
\renewcommand{\arraystretch}{1}
\caption{Cross-Database Evaluation (both SRCC and PLCC)of our proposed models and Other FR Models on SCID.}\label{tab:aStrangeTable}
\centering
\scriptsize
\renewcommand{\multirowsetup}{\centering}
\begin{tabular}{cccccccccc}
\toprule[1pt]
\multirow{2}{*}{\tabincell{c}{\textbf{\emph{Method}}}}&\multirow{2}{*}{\tabincell{c}{Distortions}}&\multirow{2}{*}{\tabincell{c}{PSNR}}
&\multirow{2}{*}{\tabincell{c}{SSIM}}
&\multirow{2}{*}{\tabincell{c}{GMSD}}
&\multirow{2}{*}{\tabincell{c}{VSI}}
&\multirow{2}{*}{\tabincell{c}{ESIM}}
&\multirow{2}{*}{\tabincell{c}{PICNN}}
&\multirow{2}{*}{\tabincell{c}{QODCNN-NR}}
&\multirow{2}{*}{\tabincell{c}{QODCNN-FR}}\cr\\
\cmidrule[0.5pt]{1-10}
\centering{\multirow{7}{*}{PLCC}}
&GN&0.955&0.936&0.954&\textbf{0.958}&\textbf{0.956}&-&0.949&\textbf{0.960}\\
&GB&0.778&\textbf{0.871}&0.797&0.836&\textbf{0.870}&-&0.845&\textbf{0.866}\\
&MB&0.763&\textbf{0.880}&0.834&0.827&\textbf{0.882}&-&0.812&\textbf{0.849}\\
&CC&0.755&0.708&\textbf{0.811}&\textbf{0.878}&0.791&-&0.752&\textbf{0.817}\\
&JC&0.839&0.859&0.935&0.915&\textbf{0.942}&-&\textbf{0.935}&\textbf{0.942}\\
&J2C&0.918&0.859&\textbf{0.943}&\textbf{0.946}&\textbf{0.946}&-&0.890&0.940\\
&Overall&0.716&0.747&\textbf{0.851}&0.697&-&0.827&\textbf{0.849}&\textbf{0.882}\\ \cmidrule[0.5pt]{1-10}

\centering{\multirow{7}{*}{SRCC}}
&GN&0.944&0.917&0.934&\textbf{0.946}&\textbf{0.946}&-&\textbf{0.947}&0.938\\
&GB&0.776&\textbf{0.870}&0.799&0.822&\textbf{0.870}&-&0.829&\textbf{0.856}\\
&MB&0.756&\textbf{0.859}&0.815&0.801&\textbf{0.861}&-&0.803&\textbf{0.828}\\
&CC&\textbf{0.732}&0.679&\textbf{0.715}&\textbf{0.816}&0.618&-&0.569&0.687\\
&JC&0.833&0.850&\textbf{0.934}&0.914&\textbf{0.946}&-&0.929&\textbf{0.935}\\
&J2C&0.907&0.846&\textbf{0.928}&\textbf{0.931}&\textbf{0.936}&-&0.865&0.916\\
&Overall&0.673&0.716&\textbf{0.843}&0.668&-&0.822&\textbf{0.848}&\textbf{0.876}\\

\bottomrule[1pt]
\end{tabular}
\end{table*}
\subsubsection{Full-Reference Image Quality Assessment}
The results of QODCNN-FR models are shown in Table II, compared with other state-of-the-art FR models: PSNR, SSIM [1], GMSD [6], SPQA [17], ESIM [18], SQMS [20],GFM [21], SFUW [19], MDOGS [22] and CNN-SQE [33]. Specially the last seven models are designed for SCIs. It can be seen from Table II that FR methods designed for SCIs achieve higher performance than FR metrics designed for NIs. The reason is that these SCI models (SPQA, ESIM, SQMS, GFM, SFUW, MDOGS and CNN-SQE) consider the correlation between HVS and local area consisting of pictorial and textual regions compared to FR-IQA models of NIs.

Among all FR metrics, QODCNN-FR can obtain the best performance and achieve a great improvement. The SRCC value of QODCNN-FR model is 2.44\% higher than the MDOGS model, 1.26\% higher than the CNN-SQE model, and the PLCC value of QODCNN-FR model is 2.32\% higher than the SFUW model, 1.02\% higher than the CNN-SQE model. In addition, our FR models fully utilize strong extracting features ability and generalization ability of CNN while CNN-SQE model only uses CNN to distinguish pictorial and textual regions which is mainly based on traditional ways.

\subsubsection{No-Reference Image Quality Assessment}
To demonstrate the excellent performance of our proposed QODCNN-NR models, it is compared with the above excellent FR-IQA models and the following state-of-the-art NR perceptual quality evaluation methods: BLINDS-II [11], BRISQUE [12], BLIQUP-SCI [23], NRLT [24], CNN-Kang [25], WaDIQaM-NR [32] and PICNN[35]. Among these NR-IQA approaches, the BLIQUP-SCI, NRLT and PICNN are designed for IQA of SCIs. In addition, CNN-SQE, CNN-Kang, WaDIQaM-NR and PICNN are CNN-based methods. For NR models, NRLT shows excellent performance among traditional NR methods, utilizing the global scope statistical luminance and texture features.

Our proposed QODCNN-NR model achieves the best performance on visual quality evaluation of SCIs compared with traditional FR and NR methods. Compared with CNN-based methods, our NR model obtains better performance than CNN-Kang model, WaDIQaM-NR model and PICNN model, and shows very close performance with CNN-SQE FR model. The PLCC value of our model is 5.66$\%$ higher than the NRLT model of NR-IQA, 1.69$\%$ higher than the MDOGS model of FR-IQA, 5.21$\%$ higher than the CNN-Kang NR model, 4.14$\%$ higher than the WaDIQaM-NR model and 0.48$\%$ higher than the PICNN model. Although the performance of PICNN is closed to the performance of our QODCNN-NR for that two models both are designed for SCIs, the proposed QODCNN-NR achieves better generalization ability from the results of cross-database experiments in Table IV. The main reason is that our method fully utilizes the strong extracting features ability of CNN, and considers the divergence of local quality within a large SCI. In addition, our method fuses local quality to obtain visual quality of images by using an adaptive and effective weighting method.

\begin{table*}[!t]
\centering
\setlength{\extrarowheight}{1mm}
\caption{Performance of QODCNN-NR-II with Different Proportion Training Data}\label{tab:aStrangeTable}
\begin{tabular}{>{\hfil}p{0.8cm}<{\hfil}>{\hfil}p{1.4cm}>{\hfil}p{0.8cm}<{\hfil}<{\hfil}>{\hfil}p{0.8cm}<{\hfil}>{\hfil}p{0.8cm}<{\hfil}>{\hfil}p{0.8cm}<{\hfil}>{\hfil}p{0.8cm}<{\hfil}
>{\hfil}p{0.8cm}<{\hfil}>{\hfil}p{0.8cm}<{\hfil}>{\hfil}p{0.8cm}<{\hfil}}
\toprule[1pt]
\multirow{2}{*}{\centering{\textbf{\emph{Index}}}} &
\multirow{2}{1.2cm}{\centering{QODCNN\\-NR-I}}&
\multicolumn{8}{c}{QODCNN-NR-II} \\
\cmidrule[0.5pt]{3-10}
~&~&80\%  &70\% &60\% &50\% &40\% &30\% &20\% &10\%\\
\midrule[0.5pt]
PLCC& 0.8849& 0.8854& \textbf{0.8908} & 0.8882 & 0.8858 & 0.8863  & 0.8877 & 0.8853 & 0.8830\\
SRCC& 0.8650& 0.8656& \textbf{0.8706} & 0.8695 & 0.8671 & 0.8678  & 0.8694 & 0.8640 & 0.8647\\
RMSE& 6.6930& 6.6697& 6.5924 & \textbf{6.5907} & 6.6426 & 6.6525  & 6.5936 & 6.6901 & 6.7400\\
\bottomrule[1pt]
\end{tabular}
\end{table*}

\subsubsection{Performance Comparison on Individual Distortion}
The performances of our models and other FR models on each individual distortion type are shown in Table III. From Table III, it can be observed that GMSD, ESIM, SFUW and MDOGS show better performance on individual distortion type compared to other traditional methods. The main reason is that these approaches consider the edge and gradient information for the visual quality prediction of SCIs. The CNN-based learning models demonstrate superiority compared to traditional methods, our FR model demonstrates excellent performance for all distortion types, and even the QODCNN-NR model outperforms traditional FR approaches and achieves competitive performance for most distortion types. Especially, the proposed NR and FR models show outstanding performance on the GN, CC, JC and LSC.
\subsection{Cross-Database Evaluation}
To verify the generalization of proposed learning models, the CNN models are trained on SIQAD and tested on SCID. Considering that two databases contain different distortion types, experimental results of our models are given on 6 common distortion types consisting of GN, GB, MB, CC, JC and J2C. All distorted SCIs of six distortion types in SIQAD are used as training set. In the procedure of testing, the common practice of Mittal \emph{et al.} \cite{mittala.{2012}} and Ye \emph{et al.} \cite{yep.{2014}} is employed, and 80\% distorted SCIs associated with 32 reference SCIs of SCID are randomly chosen to evaluate the parameters of nonlinear function (Eq. 13). The rest 20\% distorted SCIs are utilized for testing. This operation is repeated with 1000 times, and the median performance is reported.

The cross-data performance of our proposed models including NR and FR is compared with the following FR methods and NR CNN-based model: PSNR, SSIM [1], GMSD [6], VSI [5], ESIM [18] and PICNN [35]. From Table IV, it can be observed that our models achieve better performance than PSNR, SSIM, GMSD, VSI and PICNN, and our FR model obtains similar performance with ESIM designed for SCIs. Comparing the two CNN-based NR model including the proposed QODCNN-NR model and PICNN, it can be find that our NR model obtains significant performance
improvement which means the proposed model owns stronger generalization ability.

\subsection{Performance Analysis}

\begin{figure}[t]
\centering
\includegraphics[width=8cm,height=4.8cm]{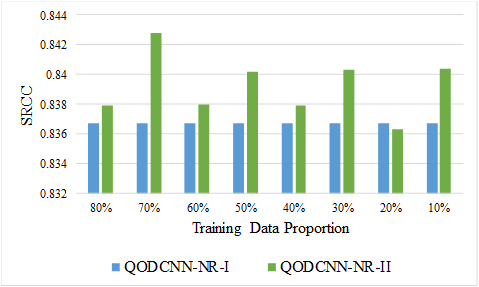}
\caption{Comparison of prediction performance under different proportion training data.}
\end{figure}


\begin{figure}[t]
\centering
\includegraphics[width=8cm,height=4.8cm]{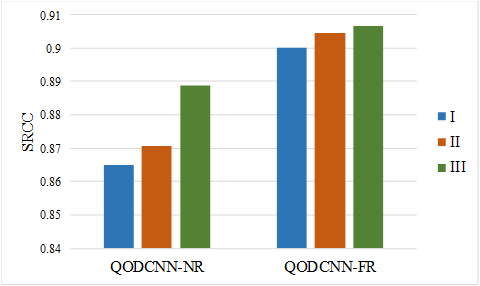}
\caption{Comparison of prediction performance in three stages of both FR and NR models on SIQAD.}
\end{figure}
\begin{figure}[t]
\centering
\includegraphics[width=8cm,height=4.8cm]{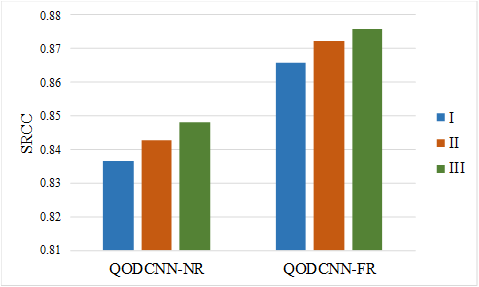}
\caption{Comparison of prediction cross-database performance in three stages of both FR and NR models on SCID.}
\end{figure}

In order to fully demonstrate the effectiveness of our optimization approaches including TMED and WLQVLSD, the pre-trained models using average weighting method of NR-IQA and FR-IQA in the first training stage are named as QODCNN-NR-I and QODCNN-FR-I, the fine-tuning models using average weighting method of second stage are denoted as QODCNN-NR-II and QODCNN-FR-II, and the fine-tuning models employing VLSD weighting method are named as QODCNN-NR-III and QODCNN-FR-III.

\subsubsection{Influence of Training Data Selection Proportion}
to choose the appropriate proportion of training data, the proportions from 80\% to 10\% are taken into account for NR-IQA. The experimental results on SIQAD are reported in Table V, and the cross-data experimental results are shown in Fig. 7. From Table V and Fig. 7, it can be seen that QODCNN-NR-II achieve better performance than QODCNN-NR-I except when the proportions of 20\% are applied. Especially, QODCNN-NR-II obtains the best performance when 70\% training data are employed to fine-tuned pretrained model. Considering that the local quality distribution is similar between FR and NR models, 70\% of the training data are employed for the second stage of FR-IQA models.

\subsubsection{Effectiveness of Two Optimizations}

To show the advantage of our two optimizations, the performance of models in three stages are shown in Fig. 8 and Fig. 9. It can be observed that the performance is improved when TMED and WLQVLSD are employed for both FR and NR models on two databases. Therefore, experimental results show that the two optimizations including TMED and WLQVLSD are effective for visual perceptual prediction of SCIs. In addition, NR model has more performance improvement compared with FR model. We consider that FR CNN-based model predicts more precise quality of image quality for utilizing reference information. TMED and WLQVLSD are two optimizations to improve accuracy of the model. When the accuracy of the model is higher, the amplitude of performance improvement is smaller.

\subsubsection{Reduce Overfitting}
One of the important problems in machine learning is overfitting. In our model, two ways are adopted to solve this problem. Firstly, data augmentation is employed to generate large data by cropping image to image patches with small size. Then in our model architecture, BN layers are added to improve learning ability of neural networks, $L_2$ regularization are used to generate sparse model, and both of them are effective approaches to reduce overfitting problem. Two regularization ways are tested to be valid and certainly helps to improve generalization ability. Our experiments on SIQAD and SCID verifies that our models achieve excellent performance on visual quality evaluation of SCIs.

\section{Conclusion}
In this paper, a neural network-based model is presented for full-reference and no-reference image quality assessment of SCIs with two optimizations. The OQDCNN model consists of three steps. In the first step, an effective CNN model is proposed to predict visual quality of SCIs for both FR and NR by employing concatenate layer to control input of reference information. Then, the Euclidean distance between DMOSs and predicted scores is employed to select more effective data, and pretrained model is fine-tuned with these data is utilized to optimize model. For third step, local weights are measured using a noise robust index VLSD and applied to fuse local quality for obtaining image visual quality. Experimental results on SIQAD demonstrate that the proposed FR and NR models achieve the best performance compared to the state-of-the-art approaches. In addition, cross-data experimental results on SCID  illustrate the strong generalization ability of our models and the efficiency of two optimizations including TMED and WLQVLSD.


%

\ifCLASSOPTIONcaptionsoff
  \newpage
\fi

\end{document}